\documentclass[10pt,journal,cspaper,compsoc]{IEEEtran}

\usepackage{graphicx,url}

\begin{document}

\title{The Preference Learning Toolbox}

\author{Vincent E. Farrugia, H\'ector P. Mart\'inez, Georgios N. Yannakakis
\IEEEcompsocitemizethanks{\IEEEcompsocthanksitem V. E. Farrugia, H. P. Mart\'inez and G. N. Yannakakis are with the Institute of Digital Games, University of Malta, Msida, 2080.\protect\\
\{vinvent.farrugia,hector.p.martinez,georgios.yannakakis\}@um.edu.mt}\thanks{}}



\maketitle

\begin{abstract}
Preference learning (PL) is a core area of machine learning that handles datasets with ordinal relations. As the number of generated data of ordinal nature is increasing, the importance and role of the PL field becomes central within machine learning research and practice. This paper introduces an open source, scalable, efficient and accessible preference learning toolbox that supports the key phases of the data training process incorporating various popular data preprocessing, feature selection and preference learning methods.
\end{abstract}


\section{Introduction}

Preferences define a central concept in human decision making and have, thus, recently inspired the development of preference-centered computational systems. Preference data has been generated from a increasing number of disciplines that investigate preference handling and preference learning\footnote{See \url{http://preferencehandling.free.fr}  and  \url{http://www.preference-learning.org/}} (PL) including artificial intelligence and machine learning (ML), decision support systems, marketing research, operations research, economics and human computer interaction. Given the increasing importance and availability of preference data, the several benefits of data collection protocols based on preferences and ranks \cite{LNCS69740437} and the advantages of PL over any other machine learning methods when ordinal data is treated \cite{martinez2014don}, preference learning \cite{furnkranz2010preference} nowadays defines one the key machine learning subareas. To facilitate the broad use of PL algorithms and methods among ML researchers, students, educators and practitioners this paper introduces the open source \emph{Preference Learning Toolbox} (PLT)\footnote{PLT is available at: \url{https://sourceforge.net/projects/pl-toolbox/}}. 

While efficient PL applications already exist they focus only on a single PL method, thereby constraining the breadth of available PL algorithms. The large variety of classification methods available in the WEKA toolkit \cite{hall2009weka} are applicable to ordinal regression tasks through the framework LPCforSOS by F\"urnkranz\footnote{Available at: \url{https://sourceforge.net/projects/lpcforsos/}}; however, this framework cannot train models from partial orders given as comparisons or non-absolute ranks (e.g., as those found in human preference reports, where measurement scales are subjective and vary across annotators and time) as it requires specialized PL methods not implemented in WEKA. In addition, that framework, as most PL applications, offers only a command-line interface making its use cumbersome without considerable prior knowledge about the PL paradigm, the particular methods and the tool itself.

PLT, instead, aims to provide an efficient collection of specialized PL methods but, most importantly, a unified and accessible user interface for them. Beyond all, PLT offers a cross-platform preference learning framework for machine learning research in order to support expert ML users but also users unfamiliar with PL methods. The toolbox includes in-line help with detailed and comprehensive documentation of the algorithms implemented and their adjustable parameters, and references to the methods used. PLT implements the key phases of a machine learning or data mining process including data preprocessing, automatic feature selection and model training. Finally, PLT is available under the GNU Lesser General Public License and offers an accessible back-end to allow for a straightforward addition of new algorithms and methods at all phases of preference learning.

\section{A Toolbox for Preference Learning: Specifications and Key Modules}

The functionality offered by PLT allows users to generate computational models that map a set of input features to an ordinal output. The process of creating a model with PLT consists of a sequence of 5 phases, each encapsulated within distinct modules to facilitate the extension of the tool. 
For the same reason, a graphical-user interface for all these phases has been implemented as separate modules decoupled from the functionality. PLT is written in Java 7 and the graphical interface is realized using JavaFX (requires Java SE 7u13 or later) which makes the tool available for a large number of Mac OS X, Linux and Windows distributions.
\\
\\
\noindent {\bf Dataset Loading and Parsing:} All preference learning techniques strive to create predictive preference models via empirical data. Naturally, the first phase of PL process involves the task of loading a dataset. In preference learning, models are constructed from a set of objects and the order (preferences) among them. PLT accepts data in single-file and dual-file format. The single format supports datasets in which a total order among the objects is known (typically given as ranks or ratings) while the dual format is tailored for datasets with partial orders (given as comparisons between two or more objects). 

Users are informed about the validity of the dataset files they have provided through UI notifications, and provide several parser parameters (e.g. symbol used to separate entries and possibility of skipping lines) to ease the process of loading diverse datasets. Currently, PLT features five example datasets users can experiment with: three synthetic datasets with known preference functions (linear, quadratic and non-linear), the popular Sushi preference dataset \cite{kamishima2011survey} and the Maze-Ball affective preferences dataset \cite{martinez2013deep}.
\\
\\
{\bf Dataset Preprocessing:} When a ML method is coupled with a particular dataset a data processing phase is usually required prior to the learning phase. 
PLT contains functionality to preview the object features within a loaded dataset, to select or omit particular features from the dataset, to alter feature representations (nominal to a binary representation and numeric to nominal) and to normalize the values of particular features (currently \emph{min-max} and \emph{z-score} normalization methods are supported).
\\
\\
{\bf Automatic Feature Selection:} While training can involve all available features, removing those which are not relevant to the sought predictive model is considered a good ML practice. In PLT, any features considered irrelevant can be deactivated manually. However, manually identifying these features is not always a trivial task while exhaustively exploring all possible feature subsets is not computationally plausible in datasets that contain many features. In such circumstances, users can opt to use an automatic feature selection (FS) algorithm. 
PLT currently supports two popular FS mechanisms, namely N-best individuals and sequential forward feature selection. Feature subsets are scored using the prediction accuracy of a model trained on the data using the PL methods described on the next phase.
\\
\\
{\bf Preference Learning Methods:} Once the dataset has been loaded and features have been preprocessed and selected, the next step is to train the model via PL. PLT enables the selection of the algorithm and its parameters as well as the validation mechanism to be used. Currently, PLT supports the ranking support vector machine method (\emph{Ranking SVM}) introduced by Joachims \cite{joachims2002optimizing} and two artificial neural network (ANN) training algorithms: backpropagation and neuroevolution. PLT's Ranking SVM module was constructed on a LIBSVM foundation \cite{chang2011libsvm}. PL performance can be assessed via k-fold cross validation and on the training set.

The current version of PLT focuses on methods that train PL models from pairwise preferences derived from total or partial orders. Note that any totally or partially ordered set can be represented as a set of pairwise preferences without any loss of information \cite{furnkranz2010preference}.
\\
\\
{\bf Reporting:} During experiment execution a console window provides the user with status updates. After each completed learning trial PLT displays a configuration summary for each phase, a section showing the accuracy values of generated models and an average method accuracy. Trained PL models can be selected and stored in JSON format.

\section{Benchmark Testing}

We evaluate the tool on synthetic and real datasets to demonstrate its efficiency and correctness. We utilise three synthetic datasets to compare the prediction accuracies and CPU times of the methods in PLT against the $SVM^{rank}$  tool \cite{joachims2002optimizing}. These datasets contain 10000 pairs of 10-feature objects that are generated using a uniform random generator;  a preference is generated for each pair, and the order is determined using a function of the object features. Each dataset uses a different function: linear, quadratic and non-linear (an artificial neural network with a hidden layer), respectively. Algorithm parameters were systematically tuned and the highest prediction accuracies are reported. The results are evaluated using 3-fold cross-validation to keep a low number of runs.

This test (see Table~\ref{tab:results}) yielded equivalent validation accuracies but significant differences in CPU times. While the linear SVM in $SVM^{rank}$ is more than twice as fast as in PLT, using a radial basis function kernel (for the non-linear dataset) or a polynomial kernel (for the quadratic dataset) yielded faster training processes in PLT. Backpropagation managed to lower the CPU times in all experiments, while neuroevvolution (as expected) tends to have a longer training time.

\begin{table*}[t!]
\centering
\begin{tabular}{ l || c | c | c }
\hline
&Linear preferences&Quadratic preferences&Non-linear preferences\\
\hline

PLT-SVM&94.41\% (00:02:48)&72.15\% (00:08:30)&85.99\% (00:07:35)\\
PLT-BP&94.94\% (00:00:13)&84.30\% (00:01:30)&84.08\% (00:01:24)\\
PLT-NE&92.15\% (00:00:51)&85.23\% (01:44:11)&85.87\% (02:00:38)\\
\hline
$SVM^{rank}$&94.82\% (00:01:33)& 77.60\% (00:46:10)&87.99\% (00:31:03)\\

\hline
  \end{tabular}
\caption{Validation accuracy and CPU time of $SVM^{rank}$ and the PLT methods (support vector machines, backpropagation and neuroevolution) on synthetic datasets. \label{tab:results}} 
\end{table*}

In addition to the tests on synthetic data, we also evaluate the correctness of the methods in PLT using the sushi preference dataset described in \cite{kamishima2010nantonac}. This dataset contains 1000 different objects with 18 features and 5000 partial orders, each with a rank of 10 objects. Following the experimental protocol described in  \cite{kamishima2010nantonac}, we evaluate the methods using  5-fold cross validation and the Spearman rank correlation coefficient. The values achieved by the PLT methods (0.377, 0.373 and 0.363 for backpropagation, neuroevolution and SVM, respectively)  are within the range of performances ([0.370,0.405]) reported by \cite{kamishima2010nantonac},  further indicating the correctness of the implementation.

\section{Conclusions and Future Work}

This paper introduced the Preference Learning Toolbox, a user-friendly application for modeling ordinal relations. 
The tool supports the key phases of any machine learning process offering preprocessing and automatic feature selection functionalities. Currently, the toolbox allows users to train SVM and artificial neural network models (via both backpropagation and artificial evolution) on input features selected with local search algorithms. Nevertheless, the tool has been designed with extensibility in mind and future work will focus on including other PL methods such as Bayesian networks \cite{chu2005preference} as well as developing PL versions of other popular methods such as hidden Markov-models.

PLT is in active development and strives to support the needs of users and developers alike. The software website provides the former with easy access to the latest distribution and updates, and allow them to suggest improvements and new functionalities. The same site offers a wiki, bug reporting capabilities and version control (git) for developers interested to partake in the improvement of the tool.

  \section*{Acknowledgment}
Special thanks go to Luca Querella for his contributions to the design and implementation of the toolbox. This work has been supported, in part, by the ILearnRW (project no: 318803) and the C2Learn (project no. 318480) FP7 ICT EU projects.

\bibliographystyle{IEEEtran}

\bibliography{bib}

\end{document}